\documentclass[10pt,conference]{IEEEtran}

\usepackage[T1]{fontenc}
\usepackage[utf8]{inputenc}
\usepackage[numbers,sort&compress]{natbib}
\usepackage{amsmath,amssymb}
\usepackage{amsthm}

\theoremstyle{remark}

\usepackage{booktabs}
\usepackage{multirow}
\usepackage{graphicx}
\usepackage{enumitem}
\usepackage{microtype}
\usepackage{caption}
\usepackage{subcaption}
\usepackage{mathtools}
\usepackage{tikz}
\usetikzlibrary{arrows.meta,positioning,fit,backgrounds}
\usepackage{fancyhdr}
\usepackage[colorlinks=true,
            linkcolor=blue!60!black,
            citecolor=blue!60!black,
            urlcolor=blue!60!black]{hyperref}
\usepackage{xcolor}

\newcommand{\codex}{\texttt{codex-py}}
\newcommand{\tq}{\textsc{TurboQuant}}

\newcommand{\nano}{\texttt{gpt-5.4-nano}}

\title{\textsc{Harbor}: Automated Harness Optimization\\
       {\normalsize A constrained-noisy-BO formulation, a reference
       algorithm, and a production case study}}

\author{\IEEEauthorblockN{Biswa Sengupta \quad Jinhua Wang}
\IEEEauthorblockA{LLM Suite Team, JP Morgan Chase \& Co.\\
Email: \{biswa.sengupta, jinhua.wang\}@jpmorgan.com}}

\begin{document}

\maketitle

% =========================================================================
\begin{abstract}
Long-horizon language-model agents are dominated, in lines of code
and in operational complexity, not by their underlying model but by
the \emph{harness} that wraps it: context compaction, tool caching,
semantic memory, trajectory reuse, speculative tool prediction, and
the glue that binds the model to a sandboxed execution
environment. We argue that harness design is a first-class
machine-learning problem and that automated configuration search
dominates manual stacking once the flag space exceeds a handful of
bits. We defend this claim in two steps. First, we formalize
\emph{automated harness optimization} as constrained noisy
Bayesian optimization over a mixed-variable, cost-heterogeneous
configuration space with cold-start-corrected rewards and a
posterior chance-constrained safety check, and give a reference
solver, \textsc{Harbor} (Harness Axis-aligned Regularized Bayesian
Optimization Routine), built from a block-additive SAAS surrogate,
multi-fidelity cost-aware acquisition, and TuRBO trust regions.
Second, we instantiate the problem in a flag-gated harness over a
production coding agent and report a controlled
four-round manual-tuning case study against a fixed task suite and an
end-to-end \textsc{Harbor} run. The formulation itself
is task-class agnostic: the configuration space, reward
correction, acquisition, and safety check apply to any agent
harness with a bounded flag space and a reproducible task suite.
\end{abstract}

\noindent{\footnotesize\itshape\textbf{Disclaimer:} This paper was
prepared for informational purposes by the LLM Suite group of JP
Morgan Chase and its affiliates (`JPMC') and is not a product of
the Research Department of JP Morgan. JP Morgan makes no
representation, warranty or undertaking whatsoever and disclaims
all liability for the completeness, accuracy or reliability of the
information contained herein. This document is not intended as
investment research or investment advice, or a recommendation,
offer or solicitation for the purchase or sale of any security,
financial instrument, financial product or service, or to be used
in any way for evaluating the merits of participating in any
transaction, and shall not constitute a solicitation under any
jurisdiction or to any person, if such solicitation under such
jurisdiction or to such person would be unlawful.\par}

% ---- Page footer: copyright line on every page --------------------------
\fancypagestyle{jpmc}{%
  \fancyhf{}%
  \renewcommand{\headrulewidth}{0pt}%
  \renewcommand{\footrulewidth}{0pt}%
  \fancyfoot[C]{\footnotesize\copyright{} 2026 JP Morgan Chase \& Co. All rights reserved \quad\thepage}%
}
\pagestyle{jpmc}
\thispagestyle{jpmc}

% =========================================================================
\section{Introduction}
\label{sec:intro}

The last eighteen months have seen a qualitative shift in how large
language models (LLMs) are deployed for long-horizon work. The
archetypal 2024 system---a prompt, a single model call, a returned
string---has been displaced by ``Agents~2.0'' or \emph{Deep Agent}
stacks that decouple planning from execution, externalize memory,
and delegate to specialized
sub-agents~\citep{agentHarnesses2026}. In this regime most of the
engineering is neither the model nor the prompt: it is the
\textbf{harness}, the deterministic code that stores, retrieves, and
serves context to the model, enforces tool semantics on its outputs,
and shields it from the failure modes that destroy long-horizon
runs~\citep{harnessreport2026}. A systematic audit of Claude
Code~\citep{diveClaudeCode} estimates that $\approx$98.4\% of its
codebase is harness---permission gates, multi-stage context
compaction pipelines, tool routing, failure recovery---and only
$\approx$1.6\% is the AI decision logic itself. A 100-tool agent
that does not aggressively manage tool descriptions can squander up
to 40\% of its context window before processing a single user
instruction~\citep{harnessreport2026}; the same harness must defend
against \emph{context poisoning}, where a hallucinated variable
becomes a self-reinforcing premise across hundreds of turns.

This paper is about the \emph{optimization} problem that the
harness creates. Teams today tune harnesses by hand: they add a
cache, run an ablation, read a few pass-rate tables, and ship or
revert. We report on doing exactly this in a real system---
\codex{}~\citep{wang2026codex}, a production Python port of the
OpenAI Codex CLI that, on Terminal-Bench, reaches parity with its
648K-LOC Rust original in just 52K LOC and ships an
enhancement layer of $\approx 30$ flag-gated capabilities---over four tuning rounds against Terminal-Bench~2%
~\citep{terminalbench2}. Each round (\textsc{A, B, C, D}) layers
\emph{additional} features on top of the $\approx\!30$ enhancements
already present in the base \codex{} harness; the flags we tune in
A--D are strictly above and beyond that baseline, drawn from the recent
agent-optimization literature:
Reflexion-style failure-self-%
reflection~\citep{shinn2023reflexion}; PASTE (Predictive
Agent-Side Tool Execution) speculative tool
execution~\citep{paste2026}; ACON (Agent-Context Observation
compressioN) gated observation compression~\citep{acon2025}; a Terminus-KIRA%
~\citep{terminuskira2026} inspired self-evaluation gate;
intent-canonicalized, content-addressable tool-result caching in
the style of \citet{semcaching2025}; tiered conversation
compression backed by the \tq{}~\citep{zandieh2025turboquant} vector
quantizer, which matches the Shannon distortion-rate lower bound
to within a small constant ($\approx\! 2.7$) and reaches absolute
quality neutrality at 3.5 bits per channel; a prompt-prefix-first
context layout grounded in observed Claude Code reuse
patterns~\citep{lmcache2025claudecode}; and a cross-session
experience library. Features are flag-gated so that the same model
and the same task set can be run with each subset enabled.

The results are sobering (\S\ref{sec:casestudy}). Of the four tuning
rounds, only B cleanly beat the no-harness-extension baseline,
scoring 17/89 against an all-flags-off baseline of 15/89 and an
Oracle (best-of-all-configurations union) of 81/89. Adding
self-evaluation on top of B dropped the full-suite score to
13/89 (C, $-4$); adding the full ACON+Reflexion+PASTE bundle on
top of that dropped it further to 12/89 (D, $-5$). The headline
causes are cross-cutting and unsurprising in hindsight---low-capability
self-evaluators make worse not better outputs; the ACON observation
gate corrupts the tool-result cache by storing the compressed
\emph{summary} rather than the raw output, so later retrievals reason
on truncated context; cold-start crippling of cross-session features;
and two silent integration bugs (cross-trial reflection memory was not
propagated between containers, and PASTE's next-tool predictor was
never wired into the runner). These are precisely the sort of causes
that a manually driven ablation loop is poorly equipped to find.

We take this as evidence that harness tuning has graduated from a
systems problem into a hyper-parameter optimization problem, and that
it should be treated with the same rigor~\citep{snoek2012bayes,
li2018hyperband, falkner2018bohb}. Our contributions are four:
\begin{itemize}[leftmargin=*, itemsep=1pt, topsep=2pt]
\item A characterization of our flag-gated \codex{} harness as a high-%
    dimensional flag-valued configuration space
    (\S\ref{sec:harness},\S\ref{sec:tq}), with a taxonomy of the six
    non-trivial interaction dimensions that make hand-tuning lossy.
\item A frankly negative case study showing how a well-motivated,
    literature-grounded manual tuning cadence over four rounds
    produces at most one statistically credible net win under a
    compute and time budget representative of industry practice
    (\S\ref{sec:casestudy}).
\item A formalization of \emph{Automated Harness Optimization} (AHO)
    as constrained noisy optimization over a structured,
    cost-heterogeneous configuration space, and a reference
    algorithm---\textsc{Harbor} (Harness Axis-aligned Regularized
    Bayesian Optimization Routine)---that combines
    Bayesian optimization over the flag bundle with a
    per-task-difficulty meta-controller and a cold-start-aware
    evaluator
    (\S\ref{sec:framework}).
\item End-to-end measurements against a fixed task suite
    (\S\ref{sec:results}), together with the harness-level
    telemetry counters that make the warm-start-aware evaluator
    (Eq.~\ref{eq:warm}) computable in practice. Concrete
    model/benchmark identities are deferred to
    \S\ref{sec:setup}.
\end{itemize}

\begin{figure}[t]
\centering
\scriptsize
\begin{tikzpicture}[
  >={Latex[length=3pt,width=3pt]},
  every node/.style={font=\scriptsize},
  box/.style={draw, rounded corners=2pt, align=center,
              inner sep=3pt, minimum height=16pt},
  inp/.style ={box, fill=blue!8},
  srg/.style ={box, fill=green!10},
  acq/.style ={box, fill=orange!10},
  det/.style ={box, fill=red!8},
  outb/.style={box, fill=gray!10}
]
\node[inp] (inX)  {Flag $\times$ fidelity\\ $(x,m)\!\in\!\mathcal{X}\!\times\!\mathcal{M}$};
\node[srg, right=10pt of inX] (surr) {SAAS\\ surrogate\\ $\mathcal{G}(\mu,\sigma^2)$};
\node[acq, right=10pt of surr] (acq)  {Cost-aware\\ acquisition\\ $\mathrm{KG}/c(m)$};
\node[outb, right=10pt of acq] (pf)    {Pareto\\ front\\ $(\mu,\mathrm{cost})$};
\node[det, below=18pt of surr] (det)  {Silent-flag\\ auto-exclusion (Axis-IV)};
\draw[->, thick] (inX) -- (surr);
\draw[->, thick] (surr) -- (acq);
\draw[->, thick] (acq) -- (pf);
\draw[->, thick] (det) -- (surr);
\draw[->, dashed] (acq.south) |- (det.east);
\draw[->, dashed] (pf.south) -- ++(0,-5pt) -| (surr.south);
\end{tikzpicture}
\caption{\textsc{Harbor} schematic. A flag-space$\times$fidelity
input $(x,m)$ is scored by a block-additive SAAS surrogate; a
cost-aware acquisition selects the next batch; runtime telemetry
(dashed) feeds Axis-IV silent-flag auto-exclusion, which prunes
dimensions from the surrogate in place. The output is a Pareto
front on (pass-rate, cost), from which a single deployment
configuration is picked subject to the posterior chance-constrained
safety check.}
\label{fig:harbor-schematic}
\end{figure}
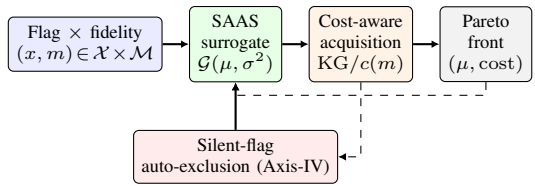

\textbf{We emphasize that our claims are empirical and ``in the small''; a
single-benchmark, single-agent ablation is the right grain for the
engineering loop, but not for conference-grade claims of universality.
Our purpose is to motivate automation, not to declare any single
configuration optimal.} Although we use a coding benchmark
(Terminal-Bench~2) as a concrete reference and \codex{} as the
harness under test, nothing in the formulation of
\S\ref{sec:framework} is specific to code: the same
flag-valued configuration space, cold-start-corrected reward,
cost-aware acquisition, and posterior chance-constrained safety
check apply to any agent-orchestration harness with a bounded flag
space and a reproducible task suite---retrieval-augmented QA
agents, customer-support copilots, web-browsing agents, scientific
tool-use agents, and so on. The coding setting is a convenience
for measurement; the method is for agent orchestration in general.
Figure~\ref{fig:harbor-schematic} summarizes the mechanism the
rest of the paper defines.

\paragraph{The case for mechanics over stacking.}
A structural lesson runs through our case study and its automated
counterpart: on any given (model, benchmark) pair, net-positive
harness features are a small, class-specific subset of the
published inventory, and the expected return from stacking more
published techniques is negative once the effective flag space
saturates the per-model signal-to-noise ratio. This is not a
claim that a given published technique is wrong---Reflexion,
ACON, PASTE, Terminus-KIRA-style self-evaluation, and
intent-canonical caching each have strong empirical results at
the scale and layer for which they were designed. It is a claim
that (i)~the per-class-architecture pick
(flag-space composition) is the load-bearing design decision,
not the per-flag ON/OFF switch;
(ii)~component-internal tuning---thresholds, fidelity
ladders, retrieval top-$k$, cache-key canonicalization---is
where most of the headline improvement lives;
and (iii)~the optimizer's job is to surface that structure
mechanically, not to certify whether a technique is
``good.'' The rest of this paper is organised accordingly: we
define the mechanics, then show what they recover on a
concrete suite, rather than arguing for or against any
individual published harness feature.

% =========================================================================
\section{Related Work}
\label{sec:related}

\paragraph{Harness as first-class subject.}
The industrialization of LLM coding agents has pushed harness
engineering into its own research
stack~\citep{harnessreport2026,agentHarnesses2026}. The ``Agents~1.0
$\to$ Agents~2.0'' / Deep-Agents transition~\citep{agentHarnesses2026}
reframes the harness as the component responsible for decoupling
planning from execution, maintaining externalized memory, and
delegating to specialized sub-agents. A frequently cited datum from
a systematic audit of Claude Code~\citep{diveClaudeCode} places
$\approx$98.4\% of its codebase in the deterministic harness
(permission gates, context pipelines, tool routing, failure recovery)
and only $\approx$1.6\% in model-facing decision logic. Adjacent work
has begun to decompose this harness by layer: key--value (KV) cache
disaggregation and hierarchical memory
(Pancake, Mem0, STITCH)~\citep{pancake2026}; prompt-prefix caching
as practiced in Claude Code~\citep{lmcache2025claudecode}; five-stage
context-compaction pipelines~\citep{harnessreport2026}; and AI
gateways (Kong, Bifrost, Portkey, LiteLLM) with their own token-bucket
rate-limiting and PII-masking guardrails. Our paper treats a
\emph{subset} of this surface---the high-level harness feature flag
space on top of a single API gateway---as an explicit optimization
problem, not as an architectural one.

\paragraph{Automated harness engineering.}
The closest prior work to ours is
\textbf{Meta-Harness}~\citep{metaharness2026}, an outer-loop system
in which an LLM proposer reads the source code, logs, and benchmark
scores of previously evaluated candidate harnesses and writes a new
one. Meta-Harness reports $+7.7$ absolute accuracy points on LawBench
with $4\times$ fewer context tokens against hand-engineered baselines,
and establishes the credit-assignment shift from model weights to
harness code that our paper then \emph{formalizes}. Complementary
efforts include \textbf{ACCLAIM}~\citep{acclaim2026}, which couples
an LLM agent with a compiler to iterate on generated code using
hardware profiling feedback; and Datadog's
\textbf{BitsEvolve}~\citep{bitsevolve2026}, which builds
``harness-first'' observability loops so that failing outputs are
automatically routed back to the agent for correction. These
systems evolve harness \emph{code}; our AHO framework searches the
harness \emph{configuration} surface with cost-aware Bayesian
optimization, and the two are complementary.

\paragraph{Agent benchmarks.}
Terminal-Bench~\citep{terminalbench2} and SWE-bench
Verified~\citep{jimenez2024swebench} have become the two standard
targets for long-horizon coding agents. Both are deliberately
task-independent, which we show (\S\ref{sec:casestudy}) amplifies the
cold-start penalty for any harness feature that is designed to
accumulate cross-task state. The EleutherAI LM Evaluation
Harness~\citep{eleutherLMEval} provides a widely-used tokenization-%
agnostic interface for running such benchmarks reproducibly.
The Terminus-KIRA
team~\citep{terminuskira2026} has reported 10-percentage-point
improvements from harness-only changes on these benchmarks; our B
result is consistent with the lower end of their estimates.

\paragraph{Harness components.}
The five core harness features we evaluate map onto an existing literature:
tool-result caching and fuzzy retrieval~\citep{khot2023decomposed};
Reflexion failure-reflection~\citep{shinn2023reflexion}; recursive
memory trees~\citep{sarthi2024raptor}; speculative tool
execution~\citep{paste2026}; ACON gated observation
compression~\citep{acon2025}; and tiered embedding compression via the
\tq{} quantizer~\citep{zandieh2025turboquant}. Binary-embedding and
Matryoshka-style two-stage
retrieval~\citep{kusupati2022matryoshka,huggingface2024binary} are
natural next steps on top of the current single-stage semantic cache
but are \emph{not} implemented in the harness evaluated here; we
include them only in the forward-looking discussion
(\S\ref{sec:discussion}).
Semantic caching at the \emph{gateway} layer has its own
substantial theory: the mismatch-cost framework
of~\citet{semcaching2025} proves that the oracle setting is NP-hard
and solves a Reverse Greedy relaxation, with a CUCB-SC
(Combinatorial Upper Confidence Bound for Semantic Caching)
offline learner for the partial-information regime and a
CLCB-SC-LS (Combinatorial Lower Confidence Bound for Semantic
Caching with List-Swap updates) online adaptive variant; these are the natural algorithmic cousins of the
acquisition function in \S\ref{sec:framework}. Dynamic model
routing~\citep{daao2025,skewroute2025,hybridllm2024} completes the
layer cake. Each of these components has independently justified
empirical benefits; our observation is that stacking them is not
additive.

\paragraph{Hyper-parameter optimization.}
Bayesian optimization for expensive black boxes has been studied for
more than a decade~\citep{snoek2012bayes, swersky2014freeze};
multi-fidelity~\citep{li2018hyperband} and combined
BO+bandit~\citep{falkner2018bohb} approaches handle heterogeneous
evaluation cost. We borrow directly from this literature in
\S\ref{sec:framework} and are, to our knowledge, the first to apply
it to harness \emph{configuration} (as opposed to harness code, \`a la
Meta-Harness) for LLM agents.

% =========================================================================
\section{The \codex{} Harness}
\label{sec:harness}

\codex{} is a production Python port of the OpenAI Codex
CLI~\citep{wang2026codex}. The port is a 12.3$\times$ code reduction
from the Rust original ($\approx$52K Python LOC versus $\approx$649K
Rust LOC) and is organized into six architectural layers: an
\emph{agent} layer (turn loop, tool orchestration, guardian approval,
multi-agent coordination, memory); a \emph{security} layer
(Seatbelt/Bubblewrap/seccomp sandboxing and execution-policy
evaluation); a \emph{protocol} layer (wire types, approval workflows,
model abstractions); an \emph{integration} layer (Model Context
Protocol client/server, OAuth authentication, backend API);
a \emph{presentation} layer (terminal UI and application server);
and an \emph{infrastructure} layer (persistent state, configuration,
telemetry, feature flags). The harness ships 34 tool handlers---
including shell execution, a unified file-patch tool, directory
listing, and a Model Context Protocol delegate---plus a layered
three-stage approval pipeline (policy check, guardian LLM risk
assessment, interactive prompt)~\citep{wang2026codex}. From a
configuration standpoint, we group the \emph{harness} surface exposed
to our ablations into five
layers:

\begin{itemize}[leftmargin=*, itemsep=1pt, topsep=2pt]
\item \textbf{Runtime.} A stateful turn loop owns dispatch, tool
  invocation, and exit conditions; it is the single integration
  point for all harness extensions and emits typed events consumed
  by the presentation and telemetry layers.
\item \textbf{Tool handlers.} A set of file tools (read, grep, glob,
  write, edit) plus shell and apply-patch, each with a read/write and
  concurrency declaration that the harness respects when parallelizing
  tool calls.
\item \textbf{Context management.} A multi-phase compactor applies
  microcompaction (inline stale-output stripping), snip compaction
  (token-threshold pruning), and full LLM-driven summarization, with
  post-compact file restoration and ghost
  snapshots~\citep{wang2026codex}. A flag-gated family of compaction
  policies---an ACON-style gate, \tq{}-based tail compression, an
  incremental strategy, an intent anchor, and a forked variant---sit
  on top of this substrate.
\item \textbf{Cross-session memory.} Two storage backends---a
  \tq{}-compressed index of (task embedding, hint) pairs and a
  trajectory-based experience library---persist compressed embeddings
  across sessions and retrieve few-shot hints when a new task embeds
  close to historical ones.
\item \textbf{Resilience / self-evaluation.} A categorized-retry
  resilience family is complemented by a self-evaluation gate in the
  runtime layer that fires a checklist-driven review before
  submission, in the style of Terminus-KIRA~\citep{terminuskira2026}.
\end{itemize}

Feature selection is unified by a four-tier flag-resolution scheme
(runtime override $>$ build-time compile flag $>$ environment variable
$>$ default); this follows the enhancement-layer design principle of keeping every
addition strictly opt-in, so that ``all enhancement flags off''
reproduces the Rust-parity baseline
exactly~\citep{wang2026codex}. The production port exposes
$\approx$35 enhancement flags across multi-agent orchestration,
safety (guardian), tool extensions, memory, context management,
productivity, session coherence, IDE integration, and observability.
For this paper the nine relevant flags, all defaulting off, are
five native harness feature flags (semantic tool-result cache,
cross-session memory index, tiered conversation compressor,
trajectory-replay experience library, speculative tool-prediction
learner), plus the self-evaluation gate, the ACON
observation-compression gate, the Reflexion reflections module,
and the PASTE next-tool predictor.

% =========================================================================
\section{The Harness Feature Layer over \codex{}}
\label{sec:tq}

The harness we evaluate packages five flag-gated extensions on top of
\codex{}. Two of these extensions (the cross-session memory index
and the tiered conversation compressor) use \tq{} as a
vector-compression primitive; the remaining three (semantic
tool-result cache, trajectory-replay experience library,
speculative tool-prediction learner) are \tq{}-free. Because the
quantizer is shared by the compression-heavy extensions, we
describe it first, but we emphasize that the harness is a collection
of flag-gated mechanisms, not a \tq{}-derived system.
\tq{}~\citep{zandieh2025turboquant} is, in its original form, an online
data-oblivious vector quantizer that achieves near-optimal
distortion rates for both mean-squared error and inner-product error.
Its core is a two-stage construction: first a random rotation of the
input vector induces a concentrated Beta distribution on each
coordinate, on which a per-coordinate Lloyd--Max
quantizer---obtained by solving a continuous $k$-means problem---is
MSE-optimal; second, a one-bit Quantized Johnson--Lindenstrauss (QJL)
transform is applied to the residual, yielding an unbiased inner-%
product estimator~\citep{zandieh2025turboquant}. The resulting
quantizer provably matches the information-theoretic
distortion--rate lower bound up to a small ($\approx 2.7$) constant,
and on KV-cache quantization achieves absolute quality neutrality at
$3.5$ bits per channel and marginal degradation at $2.5$
bits~\citep{zandieh2025turboquant}. Two properties make it a
natural primitive for agent harnesses: it is (i)~\emph{data-%
oblivious}, so no codebook fitting is required when embeddings drift
across sessions, and (ii)~accelerator-friendly (the quantizer is a
composition of a rotation, a scalar look-up table (LUT), and a
bitwise reduction),
so it can sit on the tool-result hot path without dominating
latency.

In our harness, \tq{} provides the compression primitive underneath
four extensions:

\begin{enumerate}[leftmargin=*, itemsep=1pt, topsep=2pt]
\item \textbf{Semantic tool-result cache (PolarCache).} A
  tool-result cache keyed both exactly and by compressed-inner-product
  similarity. Cache keys include the target file's POSIX
  modification time (\texttt{mtime}) as a secondary component, so
  that a stale entry is \emph{structurally} unreachable after the
  underlying file changes---rather than relying on a separate
  invalidation pass. An \emph{intent-canonical} hash on the primary
  component maps surface-form variants of a file-read
  (e.g.\ print-file, head, tail, pager, and structured read) onto a
  single canonical tuple. The design follows the intent-%
  canonicalization line of work that directly diagnoses the $\le$40\%
  hit-rate ceiling of string-level agent caches~\citep{semcaching2025}.
\item \textbf{Cross-session memory index (PolarMemoryIndex).} A
  cross-session index of (task embedding, hint text) pairs with a
  memory-type facet used to segregate Reflexion-style
  reflections~\citep{shinn2023reflexion} from trajectory hints. The
  structure follows the Zettelkasten-style agentic memory line of
  work that attaches LLM-generated keywords and contextual
  descriptions to each stored note.
\item \textbf{Tiered conversation compressor.} Keeps recent turns at
  full precision and \tq{}-compresses the tail; retrieval sits
  behind a fixed-token compaction trigger. The tiered decomposition
  mirrors \tq{}'s own two-stage quantization---coarse compression of
  stale content, fine retrieval over a recent window.
\item \textbf{Trajectory-replay experience library.} Records
  successful \emph{and} failing trajectories with $[\mathrm{OK}]$ /
  $[\mathrm{FAIL}]$ markers and injects them as few-shot hints when
  the current task embedding exceeds a similarity threshold against
  past tasks.
\end{enumerate}

A fifth native flag, the \textbf{speculative tool-prediction
learner (PASTE)}, is \emph{not} TQ-backed: it is a pure
pattern-matching predictor that learns explicit tuples of the form
(\textit{context}, \textit{next tool}, \textit{argument-derivation
function}, \textit{probability}) from trajectories and speculatively
fires the predicted tool when the tuple's probability crosses a
threshold, following PASTE~\citep{paste2026}. The online path
performs zero embedding or LLM calls; prediction is a dict lookup on
$(tool, status)$ n-grams over recent history. We include it in the
enhancement layer because, like the TQ-backed components, it
exploits cheap structural signals to reduce model calls on the hot
path---but the compression primitive it relies on is discrete
tuple-counting, not quantized inner products.

Beyond these five native flags we also evaluate three published-%
technique flags introduced in later rounds: \textbf{the
self-evaluation gate}, which interposes a second critic-style model
call between the agent's draft answer and final submission and either
confirms or rewrites it, in the style of
Terminus-KIRA~\citep{terminuskira2026}; \textbf{the ACON observation-%
compression gate}, which detects large, redundant tool outputs (long
logs, directory listings, JSON blobs) and substitutes a learned
compressed summary in place of the raw observation in the next model
turn~\citep{acon2025}; and \textbf{the Reflexion reflections module},
which after a failed task writes a short natural-language
``lesson'' into the cross-session memory index keyed by task
embedding, to be retrieved on subsequent runs of similar
tasks~\citep{shinn2023reflexion}.

A single session-scoped coordinator reads the flags at initialization
and wires itself into the pre-tool-call, post-tool-call, session-%
start, session-end, and context-hint pathways of the runtime layer.
Lazy-init failures in any single extension are caught and demoted
to a \emph{disabled} state while the originally requested flag is
preserved in the telemetry record. This graceful-degradation
policy serves two distinct purposes: (i)~a broken extension (for
example, a missing dependency or a malformed threshold) cannot
cascade into a full-suite failure, so one corrupted feature does
not invalidate a 89-task sweep; and (ii)~the divergence between
\emph{requested} flag and \emph{realized} flag is itself
observable---a write-side counter without a corresponding
consumer-side counter is the textbook signature of the exact
integration bug the D round missed, and the preserved flag record
is what an observability layer can flag as an invariant violation
(\S\ref{sec:discussion}).

% =========================================================================
\section{Case Study: Manual Tuning, A--D}
\label{sec:casestudy}

Four tuning rounds were performed by a dedicated human--agent
pair. Each round (i)~selected 2--4 published improvements,
(ii)~implemented them behind new flags, (iii)~ran the standard
7-way ablation (see Table~\ref{tab:configs}), and (iv)~produced a
Markdown report. We reproduce the essential numbers
in Table~\ref{tab:rounds} and discuss them.

\begin{table*}[t]
\centering
\small
\caption{Ablation configurations used across A--D. Each column
corresponds to one harness feature, toggled on or off at agent launch.
Column abbreviations: \textbf{PC}~semantic tool-result cache;
\textbf{PM}~cross-session memory index;
\textbf{TC}~tiered conversation compressor;
\textbf{TR}~trajectory-replay experience library;
\textbf{SP}~speculative tool-prediction learner;
\textbf{SE}~self-evaluation gate;
\textbf{AC}~ACON observation-compression gate;
\textbf{RF}~Reflexion reflections module;
\textbf{PT}~PASTE next-tool predictor.
``all-on'' and ``B-config'' are nested; D adds Tier-2 features on top
of the B flag set and keeps the self-evaluation gate \emph{off} after
its C regression.}
\label{tab:configs}
\begin{tabular}{lccccccccc}
\toprule
 & \multicolumn{9}{c}{Harness feature flags} \\
 \cmidrule(lr){2-10}
Config & PC & PM & TC & TR & SP & SE & AC & RF & PT \\
\midrule
baseline     & . & . & . & . & . & . & . & . & . \\
all-on       & \checkmark & \checkmark & \checkmark & \checkmark & \checkmark & \checkmark & . & . & . \\
B-config   & \checkmark & \checkmark & \checkmark & \checkmark & \checkmark & . & . & . & . \\
cache        & \checkmark & . & . & . & . & . & . & . & . \\
memory       & . & \checkmark & . & . & . & . & . & . & . \\
compress     & . & . & \checkmark & . & . & . & . & . & . \\
trajectory   & . & . & . & \checkmark & . & \checkmark & . & . & . \\
predict      & . & . & . & . & \checkmark & . & . & . & . \\
D-all-on   & \checkmark & \checkmark & \checkmark & \checkmark & \checkmark & . & \checkmark & \checkmark & \checkmark \\
\bottomrule
\end{tabular}
\vspace{-4pt}
\end{table*}

\begin{table*}[t]
\centering
\small
\caption{Full-suite Terminal-Bench~2 pass counts ($N{=}89$) across
A--D on \nano{} via the OpenAI Responses API. \emph{Baseline} is
the harness with every extension flag off. \emph{Oracle}
is the union of tasks passed by any configuration. B is the only
round that cleanly beat the all-flags-off baseline; C and D regressed
from it.}
\label{tab:rounds}
\begin{tabular}{lcc}
\toprule
Round & Pass count & $\Delta$ vs.\ B \\
\midrule
Baseline (all flags off)          & 15/89 & $-2$ \\
B (5 native flags)                & \textbf{17/89} & --- \\
C (B $+$ self-eval gate)          & 13/89 & \color{red}{$-4$} \\
D (B $+$ Tier-2: ACON,Refl,PASTE) & 12/89 & \color{red}{$-5$} \\
\midrule
Oracle (union over all configs)  & 81/89 & --- \\
\bottomrule
\end{tabular}
\vspace{-6pt}
\end{table*}

\begin{table}[t]
\centering
\small
\caption{Published-technique flag glossary used throughout
\S\ref{sec:casestudy}. Tier-1 (T1-$\cdot$) flags are low-risk
token-saving changes; Tier-2 (T2-$\cdot$) flags are
research-literature features that introduce new inference-time
behavior.}
\label{tab:tiers}
\begin{tabular}{@{}llp{0.49\linewidth}@{}}
\toprule
Tag & Name & Short description \\
\midrule
T1-1 & Intent-canonicalization & Hash normalization of surface-form
variants of a tool invocation so that semantically identical calls
collide in the cache. \\
T1-2 & Prompt-prefix-first & Stable system-prompt ordering for
automatic provider-side prompt caching
(e.g.\ Claude's prefix cache). \\
T1-3 & Self-eval gate & LLM-as-judge self-evaluation of a tool
output before acceptance. \\
T2-1 & Reflexion & Verbal self-reflection appended after failure,
in the style of \citet{shinn2023reflexion}. \\
T2-2 & ACON & Observation-compression gate that emits a compressed
summary of long tool outputs~\citep{acon2025}. \\
T2-3 & PASTE & Speculative-execution learner that predicts the next
tool call from short n-gram history~\citep{paste2026}. \\
\bottomrule
\end{tabular}
\vspace{-6pt}
\end{table}

\paragraph{A $\to$ B: the one clean win.}
Round B landed four changes: closing the compression retrieval loop
at the full-compaction boundary; cache-key normalization with
\emph{mtime-indexed secondary keys}---cache keys of the form
(canonical-tuple, \texttt{mtime}) where the POSIX modification time
is folded into the key so that any write to the underlying file
structurally invalidates the entry, rather than relying on a
separate timestamp-comparison pass; a failure-aware trajectory library with
$[\mathrm{OK}]$/$[\mathrm{FAIL}]$ markers in the spirit of
Reflexion~\citep{shinn2023reflexion}; and an adaptive hint-injection
threshold (raised from 0.30 to 0.85 for memory and from 0.50 to 0.80
for trajectory, with $k{=}1$). The cache hit rate moved from
$\sim$1\% (A) to 3.7\% session-weighted and $\sim$12\% intra-session
weighted, and the full-suite sweep improved from 15/89 (all-flags-off
baseline) to 17/89; zero errors introduced. B---the five
native extensions with all Tier-1/Tier-2 published-technique
flags off---remains the peak of the manual search.

\paragraph{B $\to$ C: the literature bites back.}
Round C added three Tier-1 improvements sourced from published
work: intent-canonicalization for the tool-result cache (T1-1),
motivated by the agent-caching diagnosis of
\citet{semcaching2025}; a prompt-prefix-first context layout for
automatic prompt caching (T1-2), motivated by the Claude Code reuse
patterns analyzed in~\citet{lmcache2025claudecode}; and a
Terminus-KIRA~\citep{terminuskira2026} style self-evaluation gate
(T1-3) with the self-evaluation gate bundled into the default configuration.
The full-suite sweep dropped from 17/89 (B) to 13/89 (C,
$\Delta{=}-4$).

Post-mortem forensics localized the regression to T1-3: the
self-evaluation gate fired on every turn it was consulted, and the
majority of its revisions \emph{corrected a passing answer into a
failing one}. Under a stronger model (Claude Sonnet, as used by the
original KIRA report) this step is reported a net win; under \nano{}
it is net negative. Canonicalization (T1-1) produced \emph{zero}
canonical cache hits across the full-suite sweep---the hand-written
rewrite rules for file-read, text-search, and directory-listing
aliases did not match the specific tool-call shape \nano{} chose to
emit. Prompt-prefix-first (T1-2) saved tokens but could not be
separately credited at the aggregate pass-rate level.

\paragraph{C $\to$ D: the bundle makes it worse.}
Round D retired the self-evaluation gate as default-off, added
ACON~\citep{acon2025} gated observation compression (T2-2),
Reflexion~\citep{shinn2023reflexion} style reflections (T2-1), and
a PASTE~\citep{paste2026} speculative-execution learner (T2-3). The
full-suite sweep returned 12/89, \emph{below} C and five tasks
below B. Against B, D gained two tasks
(\texttt{git-leak-recovery}, \texttt{portfolio-optimization}) and
regained three C had lost, but lost nine others. Root cause: the
ACON gate fired 171 times and elided ${\sim}501$\,KB, but the gate
was wired \emph{upstream} of the tool-result cache, so cached entries
stored the compressed \emph{summary} rather than the raw output;
tasks needing line-level detail (grep hit positions, diff hunks) then
reasoned on a truncated view and failed. Two latent integration bugs
compounded it and the manual loop surfaced neither: the
Reflexion-retrieval counter was identically zero (the cross-session
memory store was not propagated between container trials, so
reflections written by one container were invisible to the next), and
the PASTE-prediction counter was identically zero (the predictor's
entry-point was never actually invoked by the agent runner, so
speculative-tool execution never fired).

\paragraph{What the four rounds teach us.}
Six structural facts about manual harness tuning:
(i)~each round adds code but produces a \emph{worse} full-suite pass
rate than B from strictly less code;
(ii)~configuration space is evaluated once (Oracle $=$ 81/89 vs.\
B peak $=$ 17/89 shows 64 tasks are passable by \emph{some}
config but not by the best single one);
(iii)~feature interactions are not additive---Tier-2 on top of B
actively corrupts upstream state;
(iv)~cross-session features are cold-start-bound and in D two
consumer counters were identically zero from integration bugs the
manual loop missed;
(v)~literature lift for one model does not transfer to another
(T1-3 on \nano{} vs.\ KIRA on Claude Sonnet);
(vi)~full-suite sweeps are expensive (\S\ref{sec:results}) but are
the only evidence level at which small per-flag effects clear
noise.

% =========================================================================
\section{Automated Harness Optimization}
\label{sec:framework}

We now formalize what the A--D loop was approximating and describe
how we would replace it. Let the harness configuration space be a
product
\begin{equation}
  \mathcal{C} \;=\; \prod_{i=1}^{F} \mathcal{C}_i,
\end{equation}
where each $\mathcal{C}_i$ is either a Boolean flag, a numeric
threshold, or a discrete preset (e.g.\ ``cache.similarity$\in\{0.75,
0.80, 0.85\}$''). For the codebase in this paper, $F\approx 40$ with
$|\mathcal{C}|$ far exceeding $2^{40}$. In practice only a handful
of these dimensions carry signal on any given (model, benchmark):
our Axis-IV detector (\S\ref{sec:impl-silent}) auto-excludes
silent flags at runtime, typically collapsing the effective flag
count to $F_{\mathrm{eff}}\!\approx\!5$. The optimizer's
per-iteration work is dominated by $F_{\mathrm{eff}}$, not by $F$;
the full $F\!\approx\!40$ is the nominal ambient dimension the
SAAS prior is designed to tolerate.

Let $\mathcal{T}$ be the task suite and $R(c,t)\in\{0,1\}$ the pass
outcome of configuration $c$ on task $t$; at fixed $t$, $R$ is
Bernoulli with execution noise from LLM sampling and the network.
In our case study $\mathcal{T}$ is the fixed 89-task TB~2 suite,
so the outer expectation is a uniform sum over 89 tasks and the
stochasticity in $R(c,\cdot)$ is execution noise at fixed task
identity; a Wilson score interval~\citep{wilson1927} on the total
pass count---the standard small-sample binomial confidence interval,
obtained by inverting the score test for a Bernoulli proportion and
preferred over the normal-approximation interval at low
$p$---is therefore valid. Per-evaluation cost $\mathrm{cost}(c,t)$ is itself
configuration-dependent.

We define Automated Harness Optimization as the following
ground-truth constrained optimization, parameterized by a
\emph{deployment budget} $B_{\mathrm{dep}}$---an \emph{expected
per-task cost ceiling} (measured in compute-seconds or API dollars)
that the returned configuration $c^\star$ must respect \emph{at
inference time}, averaged over the task distribution---and a safety
margin $\delta$ relative to the measured all-flags-off baseline
pass rate $R_0$:
\begin{align}
c^\star \;=\; &\arg\max_{c\in\mathcal{C}}\ \mu(c)
\ \ \text{with}\ \ \mu(c)\equiv\mathbb{E}_{t}\!\left[R(c,t)\right]
\label{eq:aho}\\
\text{s.t.}\ \ &\mathbb{E}_{t}\!\left[\mathrm{cost}(c,t)\right]
    \le B_{\mathrm{dep}},\notag\\[-1pt]
&\mu(c)\;\ge\;R_0 - \delta.\notag
\end{align}
This is the ground-truth problem: $\mu$ and the cost are
unobservable populations, so any solver must enforce the safety
constraint \emph{algorithmically}. We do this with a posterior
chance-constraint~\citep{schreiter2023riskaverseBO},
$\Pr_{\mathrm{post}}\!\bigl(\mu(c)\ge R_0-\delta\bigr)\ge 1-\eta$,
evaluated under the surrogate's posterior at each candidate---this
is a property of the optimizer's beliefs, not of $\mu$. The solver
is additionally constrained by an outer-loop \emph{search budget}
$B_{\mathrm{sea}}$---the total compute or dollars available to the
optimizer \emph{across} all noisy evaluations during the outer
loop, i.e.\ a cap on the cumulative evaluation history
$\mathcal{H}$ rather than on any single evaluation:
$\sum_{(c_i,t_i)\in\mathcal{H}}\mathrm{cost}(c_i,t_i)\le
B_{\mathrm{sea}}$. This constraint terminates the \emph{while}-loop
in Algorithm~\ref{alg:ahosas} rather than entering the argmax over
$c$. The two budgets are distinct and play different roles:
$B_{\mathrm{sea}}$ is an amortised one-time cost paid during
optimisation, while $B_{\mathrm{dep}}$ constrains the deployed
configuration forever after. Three properties differentiate AHO from standard hyper-%
parameter optimization.

\paragraph{Property I: Cold-start-corrected reward (mixture model).}
Cross-session features require state that is empty on the first
session and partially primed on later ones, so short sweeps
underestimate fully-warm performance. Let $n$ be the number of prior
primed sessions and let
$p_{\mathrm{base}}\equiv p_{\mathrm{base}}(c_{\mathrm{cold}})$ be the
pass rate of the \emph{all-flags-off baseline}---i.e.\ the configuration
$c_{\mathrm{cold}}$ with every warm-state-dependent flag disabled
(directly measured; Table~\ref{tab:baselines},
$p_{\mathrm{base}}=15/89$ on \nano{}). We treat $p_{\mathrm{base}}$
as a scalar rather than a function of $c$: the non-warm flags we
vary (caching thresholds, compaction sizes, etc.) have negligible
effect on a run in which all warm-state-dependent features are
disabled by construction, so $p_{\mathrm{base}}(c)\approx
p_{\mathrm{base}}$ on the support of the solver's queries at $w{=}0$.
The mixture model
\begin{equation}
\!\!\mathbb{E}[R(c,t)\mid n]
\,=\, w(c,n)\,p_\infty(c) + \bigl(1{-}w(c,n)\bigr)\,p_{\mathrm{base}},
\label{eq:warm}
\end{equation}
with $w(c,n)\in[0,1]$ the warm-up \emph{fraction} after $n$ primed
sessions; inverting gives
\begin{equation}
\hat p_\infty \;=\; \bigl(\hat p_{\mathrm{obs}} - (1{-}\hat w)\,
\hat p_{\mathrm{base}}\bigr)/\hat w,
\label{eq:warminv}
\end{equation}
which---unlike a bare multiplicative correction---reduces to
$\hat p_{\mathrm{obs}}$ at $\hat w{=}1$, stays in $[0,1]$ \emph{in
expectation} under the mixture model, and is clipped to $[0,1]$
whenever finite-sample noise pushes the raw estimate outside.
At the clip boundary we inflate the observation-term contribution
of Eq.~\ref{eq:deltavar} to its worst-case Bernoulli value
$\hat w_i^{-2}/4$ (an upper bound on the $\hat\sigma^2_{\mathrm{obs}}/
\hat w_i^2$ summand only---the $\hat\sigma^2_{\mathrm{base}}$ and
$\hat\sigma^2_w$ summands are retained on top of it, not replaced by
it).
At $\hat w{=}0$ the target $p_\infty$ is unidentified from a single
$\hat p_{\mathrm{obs}}$: we degrade the point to zero-precision
evidence ($\sigma_i^2\to\infty$ floored at the prior variance) and
rely on the GP to ignore it. We compose $w$ per flag with a
pessimistic $\min$-rule
$w(c,n)=\min_{f\in c_{\mathrm{warm}}} w_f(n)$. This is a
lower bound on the system-level warm fraction under a conjunctive-%
cascade reading---the system is only as warm as its least-primed
feature---and trades bias (under-correction of warm benefit) for
robustness to unknown cross-flag coupling. Principled alternatives
(product rule under independence, saturating logistic composition)
correspond to stronger parametric assumptions about how partially
primed features combine, and remain fit-time options. Each
$w_f(n)$ is fit from logged cross-session counter trajectories
(\S\ref{sec:setup}). Letting
$\theta(\hat p_{\mathrm{obs}},\hat p_{\mathrm{base}},\hat w)=
(\hat p_{\mathrm{obs}}-(1-\hat w)\hat p_{\mathrm{base}})/\hat w$,
the delta-method variance of $\hat p_\infty$ in terms of the
independent noise of the three estimators is
\begin{equation}
\!\!\sigma_i^2
 \approx \frac{\hat\sigma^2_{\mathrm{obs},i}}{\hat w_i^2}
 + \frac{(1{-}\hat w_i)^2\,\hat\sigma^2_{\mathrm{base}}}{\hat w_i^2}
 + \frac{(\hat p_{\mathrm{obs},i}{-}\hat p_{\mathrm{base}})^2
         \,\hat\sigma^2_{w,i}}{\hat w_i^4},
\label{eq:deltavar}
\end{equation}
passed to the GP as a heteroscedastic diagonal so that cold-start
points ($\hat w\downarrow 0$) are not spuriously confident. A
calibration requirement, learned the hard way in our own runs:
$\hat p_{\mathrm{base}}$ must be measured at the \emph{same}
fidelity $m{=}N$ as the acquisition, and on the same
no-warm-state configuration, or else every first-order
coefficient $\alpha_\ell^2$ picks up a systematic negative bias.
A cheap-fidelity proxy ($m{=}10$) we initially used inflated
$\hat p_{\mathrm{base}}$ relative to the true $m{=}89$ value of
$15/89$, and the inversion of Eq.~\ref{eq:warminv} then returned
a raw $\hat p_\infty$ outside the unit interval (i.e., the
point estimate of the warm-fraction-inverted mixture fell below
zero before clipping) for almost every candidate, collapsing the
additive part of the surrogate; the cross-block residual
$\alpha_\times^2$ absorbed the signal only weakly. The fix is
to anchor $\hat p_{\mathrm{base}}$ to a single direct measurement
at the full fidelity and treat it as a scalar rather than a
surrogate-estimated quantity. Caveat: on TB~2 with \nano{}, the B
peak sits only $0.022$ above $p_{\mathrm{base}}$, so
Eq.~\ref{eq:warm} is well-specified but tells us honestly that
warm-start effects are not separable from noise on this suite.

\paragraph{Property II: Mixed-input, block-additive surrogate with residual interactions.}
Not all flags matter, and flags cluster by layer (caching, memory,
compaction, prediction, evaluation gate). Since ${\sim}35$ of the
$F{\approx}40$ dimensions are Boolean or categorical, we embed them
using the mixed-variable treatment of~\citet{daulton2022mixedbo}
(Booleans $\{0,1\}\mapsto\{-1,+1\}$; categorical presets via the
latent-relaxation of~\citet{garridomerchan2020categorical}) so that
the Mat\'ern-$5/2$ kernel is well-defined on each block. Let
$\Pi=\{\pi_\ell\}_{\ell=1}^{L}$ partition the $F$ dimensions into
$L$ layer blocks, and for each configuration $c$ let $c_{\pi_\ell}$
be its projection onto block $\ell$. Writing
$K_\ell(c,c')\equiv k^{\mathrm{M}5/2}_{\theta_\ell}(c_{\pi_\ell},
c'_{\pi_\ell})$ for the per-block Mat\'ern-$5/2$ kernel, we form the
surrogate kernel from per-block main effects and \emph{pairwise
tensor-product} cross-block interactions in the Duvenaud-additive
style~\citep{duvenaud2011additive}:
\begin{equation}
\!\!k(c,c') = \sum_{\ell=1}^{L}\!\alpha_\ell^2\,K_\ell(c,c')
\;+\; \alpha_\times^2\!\!\sum_{\ell<\ell'}\!K_\ell(c,c')\,K_{\ell'}(c,c').
\label{eq:saaskernel}
\end{equation}
Each summand is a product of PSD kernels and is therefore itself PSD
by the Schur product theorem, so $k$ is PSD by construction---no
Gram-matrix projection is required, and subtraction-style
``residual'' constructions (which are not PSD for Mat\'ern
kernels in general, since $k^{\mathrm{M}5/2}(c,c')-\sum_\ell
K_\ell(c,c)$ is negative on the diagonal whenever $L>1$) are avoided.
An axis-aligned SAAS (Sparse Axis-aligned Subspace) half-Cauchy
prior $\theta_{\ell,d}^{-2}\sim\tau_\ell\,\mathrm{HC}(1)$
(where $\mathrm{HC}(1)$ denotes the standard half-Cauchy
distribution)~\citep{eriksson2021saasbo}
contracts irrelevant per-layer dimensions, and a half-Cauchy prior
on $\alpha_\times$ tightly concentrated near zero (scale
$\tau_\times\ll\min_\ell\tau_\ell$) softly identifies the cross-block
scale: $\alpha_\times^2$ is pulled to zero unless evidence demands a
pairwise interaction such as the ACON$\leftrightarrow$cache coupling
surfaced by round D, which a strictly additive kernel would forbid
by construction. Eq.~\ref{eq:saaskernel} is a hybrid of the
Duvenaud-additive kernel with a \emph{per-block} main-effect scale
$\alpha_\ell^2$ (to accommodate SAAS-style per-block sparsity) and
a \emph{single} cross-block scale $\alpha_\times^2$ shared across
all pairs (so the number of hyper-parameters grows linearly in $L$
rather than quadratically). Higher-order (three-way and above)
cross-block interactions are excluded by this truncation, a
deliberate model choice that trades the ability to detect three-way
couplings for identification robustness. Under arbitrary
input measures the $\alpha_\ell^2$ and $\alpha_\times^2$ are not
strictly orthogonal Sobol/Hoeffding variance components, but are
weakly identified by the $\tau_\times\!\ll\!\tau_\ell$ prior and
remain coarsely interpretable as ``per-block'' vs.\ ``cross-block''
sensitivity.

\paragraph{Property III: Multi-fidelity, cost-aware, safe acquisition.}
A \emph{single} full-suite evaluation takes ${\sim}2$\,h of
wall-clock at concurrency 4 but consumes a non-trivial fraction of
the daily LLM API quota; exhaustive search over $F{\approx}40$
Boolean flags is not wall-clock-bound but API-cost-bound, and even
at the observed wall-clock the fixed daily token budget limits
full-suite acquisitions to ${\sim}10$ per week in practice. The
arithmetic is immediate: at the observed
$\approx$1.5\,M-token-per-full-suite-evaluation cost for \nano{} on
TB~2 (89 tasks $\times$ median 17K tokens per trajectory) and a
fixed organizational budget of $\approx$2.2\,M tokens per business
day on this account, the daily ceiling is
$\lfloor 2.2\text{M}/1.5\text{M}\rfloor\!=\!1$ full-suite
evaluation per day on weekdays, giving
$\approx\!5$--$10$ full-suite evaluations per
seven-day calendar week after accounting for queue backoff and
re-runs for flaky tasks. Concretely, at $B_{\mathrm{sea}}$ fixed
to one calendar week, the optimizer cannot afford more than
${\sim}10$ full-suite samples; everything else must be spent at a
cheaper fidelity. We
therefore index the
acquisition by a task-subset \emph{fidelity}
$m\in\{8,22,44,89\}$, where $\mathcal{T}_m\subset\mathcal{T}$ is a
stratified subset with $|\mathcal{T}_m|=m$ drawn once and fixed, and
use the fidelity-aware qNEHVI (q-batch Noisy Expected
Hypervolume Improvement)
of~\citet{daulton2022mixedbo,balandat2020botorch} over the two
objectives $\bigl(\hat\mu(c;\mathcal{T}_m),\,-\hat C(c;\mathcal{T}_m)\bigr)$
with $\hat\mu(c;\mathcal{T}_m)=\frac{1}{m}\sum_{t\in\mathcal{T}_m}R(c,t)$
and $\hat C(c;\mathcal{T}_m)=\frac{1}{m}\sum_{t\in\mathcal{T}_m}\mathrm{cost}(c,t)$.
Because each $R(c,t)$ is Bernoulli, $\hat\mu(c;\mathcal{T}_m)$ is a
mean of Bernoullis with intrinsic variance $\mu(c)(1-\mu(c))/m$
that depends on the input; we pass this heteroscedasticity through
the GP as an observation-noise diagonal together with Eq.~\ref{eq:deltavar},
treating the Gaussian-likelihood surrogate as the standard
BO approximation to a Binomial-likelihood observation model.
The acquisition trades information-per-dollar across fidelities
with a multi-fidelity cost denominator in the style
of~\citet{wu2020mfkg}. The ratio form is provably cost-rational
only within the Knowledge-Gradient family of acquisition functions;
applied to qNEHVI it is an engineering heuristic with the right
scaling behavior rather than a theorem.
\begin{align}
a(c_{1:q}, m) \;&=\;
\frac{q\mathrm{NEHVI}_{m}\bigl(c_{1:q};\mathcal{G}\bigr)}
     {\mathrm{Cost}(c_{1:q}, m)},
\label{eq:cost-aware}\\[-1pt]
\mathrm{Cost}(c_{1:q}, m)\;&\equiv\;
\sum_{i=1}^{q}\sum_{t\in\mathcal{T}_m}\mathrm{cost}(c_i, t).\notag
\end{align}
Here $q\mathrm{NEHVI}_m$ is evaluated against the Pareto frontier
of subsample-consistent fidelity-$m$ estimates, so incumbents from
different fidelities are not mixed. Cost enters in two distinct
roles without redundancy: as a negated \emph{objective}, it drives
Pareto-frontier exploration over configurations (the solver can
trade pass-rate for dollars on the returned $c^\star$); as the
\emph{denominator}, it weights per-dollar information gain and
selects the cheapest fidelity $m$ at which the acquisition is
justified. The first is a property of the returned solution, the
second is a property of the acquisition policy. The safety
constraint is applied \emph{per candidate} in the batch: each
$c_i$ in the qNEHVI batch must satisfy the posterior chance
constraint of~\citet{schreiter2023riskaverseBO},
$\Pr_{\mathrm{post}}(\mu(c_i){\ge}R_0{-}\delta)\ge 1{-}\eta$,
equivalent under a Gaussian surrogate posterior to
$\mu(c_i)-\Phi^{-1}(1{-}\eta)\,\sigma(c_i)\ge R_0-\delta$---i.e.,
candidates whose lower credible bound regresses the all-flags-off
baseline by more than $\delta$ are rejected, structurally blocking
the C-style ``literature bites back'' failure mode.

\paragraph{Reference algorithm: \textsc{Harbor}.}
The Sparse Axis-aligned Subspace (SAAS) prior of
\citet{eriksson2021saasbo} is the natural choice for
this problem class: the flag space $\mathcal{C}$ is nominally
high-dimensional ($F\!\approx\!40$ in our harness, $2^{F}$
corners), yet on any given (model, benchmark) pair few flags
carry signal---the rest are either silent (Axis-IV detector,
\S\ref{sec:impl-silent}) or dominated by telemetry-level
integration failures. SAAS concentrates a half-Cauchy prior
on per-dimension lengthscales near zero with a heavy tail,
so the posterior contracts irrelevant axes aggressively while
leaving a handful of active dimensions free---exactly the
shape we expect. Algorithm~\ref{alg:ahosas} couples this SAAS
block-additive surrogate with cross-block residual (P2), the
multi-fidelity cost-aware constrained batch acquisition (P3),
a trust-region local-move subroutine (TuRBO, Trust-Region
Bayesian Optimization; \citealp{eriksson2019turbo}) for
high-dimensional search, and the mixture-based warm-start
correction (P1). It is warm-started from a meta-learned
posterior when a prior harness generation's log is available
(\citealp{golovin2017vizier, ram2022warmstart}), otherwise
from a Sobol design at the cheapest fidelity.

\begin{figure}[t]
\centering
\fbox{\begin{minipage}{0.94\linewidth}
\footnotesize
\textbf{Algorithm 1: \textsc{Harbor}} --- Harness Axis-aligned
Regularized Bayesian Optimization Routine.\\[3pt]
\textbf{Input:} search budget $B_{\mathrm{sea}}$, deploy budget
$B_{\mathrm{dep}}$, task suite $\mathcal{T}$, prior model
$\mathcal{G}_0$, safety margin $\delta$, risk level $\eta$,
block-freeze count $n_{\min}$.\\[1pt]
\textbf{Output:} Pareto set of safe, budget-feasible configurations
on $(\mu, \mathrm{cost})$.\\[3pt]
\begin{tabbing}
xxx\=xxx\=xxx\=xxx\=\kill
\ 1.\>initialize GP $\mathcal{G}\!\leftarrow\!\mathcal{G}_0$ with\\
\>\ kernel Eq.~\ref{eq:saaskernel}\\
\ 2.\>$D_0\!\leftarrow\!\mathrm{Sobol}(32)$ at fidelity $m\!=\!8$;\\
\>\ evaluate and refit $\mathcal{G}$\\
\ 3.\>place $M$ trust regions $\{\mathrm{TR}_k\}$ at the\\
\>\ top-$M$ posterior-mean incumbents\\
\ 4.\>\textbf{while} history cost $<B_{\mathrm{sea}}$ \textbf{do}\\
\ 5.\>\>\textbf{for each} trust region $\mathrm{TR}$ \textbf{do}\\
\ 6.\>\>\>pick batch $(c_{1:q},\,m)$ maximizing\\
\>\>\>\ $\dfrac{q\mathrm{NEHVI}_m(c_{1:q})}{\mathrm{Cost}(c_{1:q},m)}$\\
\>\>\>\ subject to $c_i\in\mathrm{TR}$ and\\
\>\>\>\ per-$c_i$ safety constraint:\\
\>\>\>\ $\mu(c_i){-}\Phi^{-1}(1{-}\eta)\,\sigma(c_i)\!\ge\!R_0{-}\delta$\\
\ 7.\>\>\>evaluate batch on subset $\mathcal{T}_m$\\
\ 8.\>\>\>invert warm-start (Eq.~\ref{eq:warminv});\\
\>\>\>\ attach $\sigma_i^2$ (Eq.~\ref{eq:deltavar}); clip\\
\ 9.\>\>\>refit $\mathcal{G}$ by NUTS (No-U-Turn\\
\>\>\>\ Sampler) on the SAAS prior;\\
\>\>\>\ recompute $\{\alpha_\ell^2,\alpha_\times^2\}$\\
10.\>\>\>update $\mathrm{TR}$ by the TuRBO success rule\\
11.\>\>\>\textbf{if} $\alpha_\ell^2$ collapses \textbf{and}\\
\>\>\>\ $n_\ell\!\ge\!n_{\min}$ \textbf{then} freeze block $\ell$\\
12.\>\textbf{return} Pareto front on $(\mu,\mathrm{cost})$ under\\
\>\ the GP posterior, restricted to\\
\>\ $\bigl\{c:\,\mathbb{E}_t[\mathrm{cost}(c,t)]\!\le\!B_{\mathrm{dep}}$\\
\>\ \ \ and safety constraint holds$\bigr\}$
\end{tabbing}
\end{minipage}}
\caption{Reference \textsc{Harbor} solver for harness configuration.
$\mu(c),\sigma(c)$ denote GP posterior mean and standard deviation;
$\mathcal{T}_m$ is the fidelity-$m$ task subset; $\alpha_\ell^2$ is
the per-block kernel scale of Eq.~\ref{eq:saaskernel}; $n_\ell$ is
the number of distinct block-$\ell$ projections seen in history
$\mathcal{H}$.}
\label{alg:ahosas}
\end{figure}

Four design choices distinguish \textsc{Harbor} from vanilla
GP-EI (Gaussian-Process Expected Improvement).
(i)~\emph{Sparsity-by-prior} (SAAS) at $F{\approx}40$
\citep{eriksson2021saasbo}; (ii)~\emph{TuRBO locality} for
high-dimensional sample efficiency; (iii)~\emph{multi-fidelity task
subsets} amortize early exploration on cheap $m$ and escalate only
when posterior variance reduction justifies it; (iv)~\emph{posterior
chance-constrained safety} blocks C-style regressions, with a
minimum-observation guard $n_\ell\ge n_{\min}$ preventing premature
freezing. The self-evaluation gate and other model-quality-dependent flags
are treated as surrogate \emph{context variables} from a capability-%
conditioned table rather than searched.

% =========================================================================
\section{Experimental Setup}
\label{sec:setup}

\paragraph{Benchmark.} Terminal-Bench 2.0 (TB~2) comprises 89 terminal
tasks spanning shell scripting, systems administration, cryptography,
COBOL modernization, scientific-Python porting, and adversarial
diagnostics. Each task runs in its own Docker container;
a task is counted as passed iff the container's verifier awards
\texttt{reward = 1.0}. All headline numbers in this paper are reported
against the full 89-task suite.

\paragraph{Models.}
All headline numbers in this paper are measured with \nano{} served
through the OpenAI Responses API. The same flag-gated harness is
used throughout, and the benchmark task specifications are
unchanged across rounds.

\paragraph{Harness.}
The harness at the D snapshot, running on \codex{}. Harness
extensions are toggled at agent launch by a bundle of feature-flag
settings resolved through the four-tier flag-resolution scheme
described in \S\ref{sec:harness}.

\paragraph{Metric.} Unless otherwise noted, the headline metric is
\emph{pass rate} (passed tasks / total tasks). We additionally report
a small set of per-session harness telemetry counters---cache hits,
canonical cache hits, compressed turns, estimated tokens saved, ACON
bytes elided, reflections written and retrieved, and PASTE predictions
fired---averaged per configuration. These counters are emitted by the
harness as structured events at session end; they are the same
counters that feed the warm-start estimator
(Eq.~\ref{eq:warm}).

\paragraph{Implementation.}
The \textsc{Harbor} implementation evaluated in this paper is
pure-Python and substitutes three pieces of the Bayesian
scaffolding from \S\ref{sec:framework} with closed-form analogues.
(i)~A weighted ridge regression on $\{\alpha_\ell^2,\alpha_\times^2\}$
with fixed penalties $\alpha_{\mathrm{cross}}\!\gg\!\alpha_{\mathrm{main}}$
replaces SAAS-prior${+}$NUTS sampling; this preserves the
$\tau_\times\!\ll\!\tau_\ell$ contraction behaviour of the surrogate
but not the full posterior over kernel scales. (ii)~A linear
tensor-product kernel on $\pm 1$-valued Booleans replaces the
Mat\'ern-$5/2$ kernel; the two are equivalent on binary inputs up
to a rescaling, since the Mat\'ern-$5/2$ kernel on
$\{-1,+1\}^{d}$ reduces to an affine function of the Hamming
distance. (iii)~A per-candidate Monte-Carlo EHVI with a greedy
diversity term replaces the joint qNEHVI batch; this is tight at
the operating batch size $q{=}1$ used throughout and an
approximation at $q{>}1$. The stratified task subset
$\mathcal{T}_m$ is realised as a seeded-prefix shuffle drawn once
and held fixed. All other correctness-critical elements of
\S\ref{sec:framework}---Eqs.~\ref{eq:warm}--\ref{eq:cost-aware},
the posterior chance constraint, the warm-start variance
inflation (including the term-1-only inflation at the clip
boundary), and the per-block freeze rule---are implemented
literally. We flag these substitutions up-front so that the
reference specification in \S\ref{sec:framework} is not confused
with the evaluated artifact.

\paragraph{Preflight smoke verification.}
Before committing compute to any $m{=}89$ evaluation, each
\textsc{Harbor}-proposed configuration is subjected to a
$\sim$$4$-minute smoke on a single fast task
(\texttt{configure-git-webserver}) whose purpose is not to
measure pass-rate but to verify that every on-flag produces
non-zero telemetry. Flags carry heterogeneous firing
thresholds---tiered compression requires $\ge 5$ turns to
trigger, ACON $\ge 3$, cache $\ge 2$---so the check is
turn-count aware: a counter of zero is downgraded from RED
to YELLOW when the smoke trajectory was too short for the
firing threshold to be reached. Configurations with any
remaining RED flag are vetoed and the next Pareto candidate
is proposed. This pre-flight gate turns the commit criterion
from ``posterior argmax'' into ``posterior argmax conditional
on every on-flag being structurally live,'' which in
practice was the difference between picking silent
configurations and committing only runs whose telemetry
will exist.

\paragraph{Silent-flag auto-exclusion.}
\label{sec:impl-silent}
\textsc{Harbor} extends the SELF\_EVAL ``context-variable''
treatment of \S\ref{sec:casestudy} to any flag that is
empirically silent on the current (model, benchmark) pair.
After each evaluation an analyzer screens every
on-observation of flag $f$ for its warm counter
$\bar n_f$; if $\bar n_f < \varepsilon$ across
$\ge n_{\mathrm{silent}}$ on-observations (default $3$), $f$
is marked silent and pruned from the search space for
subsequent acquisitions. This is the
mechanism by which the paper's diagnosis of isolated
integration bugs becomes a standing search-space pruner.

% =========================================================================
\section{Results}
\label{sec:results}

\paragraph{Baselines.}
Table~\ref{tab:baselines} shows the all-flags-off baseline (every
native extension flag off) on the full 89-task suite for
\nano{}, the B peak (the five native flags with every
Tier-1/Tier-2 flag off; see Table~\ref{tab:configs}), and the
Oracle upper bound (union of tasks passed by any configuration we
ran). Both models share the same
\codex{} runtime, tool handlers, and sandbox; the harness switches
between providers through a single configuration toggle, so nothing
but the model changes. The A--D ablation rounds were all executed
with \nano{} because only its API quota was compatible with running
the entire sweep at the required concurrency. The gap between B
(17) and Oracle (81) is the actionable AHO signal:
most tasks are passable by \emph{some} configuration, but the
best \emph{single} configuration we found captures only 21\%~of
that union.

\begin{table}[t]
\centering
\small
\caption{\nano{} full-suite pass counts on Terminal-Bench~2.
\emph{Baseline} is the harness with every native extension flag
off; B is the manual peak (5 native flags); D is the
all-on manual stack (8 flags); \textsc{Harbor} is the automated
solver's returned configuration; \emph{Oracle} is the best-of-any-%
config union.}
\label{tab:baselines}
\begin{tabular}{lc}
\toprule
Configuration & 89-task pass count \\
\midrule
Baseline (no \tq{})              & 15 / 89 \\
B (5 native flags)         & 17 / 89 \\
D (all-on, 8 flags)              & 12 / 89 \\
\textbf{\textsc{Harbor}} (2 flags)  & \textbf{17 / 89} \\
Oracle (union over configs)      & 81 / 89 \\
\bottomrule
\end{tabular}
\vspace{-4pt}
\end{table}

\paragraph{Feature-level deltas.}
Telemetry collapsed across B--D shows three points: (i)~B cache-key
normalization (collapsing numeric read-offset and read-length arguments
onto a single key) was the real cache win; C canonicalization added
zero additional hits. (ii)~D ACON fired $171$ times and elided
${\sim}501$\,KB, but the gate was wired upstream of the tool-result
cache, so the cache stored the compressed summary rather than the raw
output---tasks needing line-level detail then reasoned on a truncated
view. (iii)~Two D counters were hard-zero: the cross-session memory
store was never propagated between container trials (reflections
written but never retrieved), and the PASTE predictor's entry-point
was never invoked (speculative predictions computed but never
executed). Both bugs would have been caught by a counter-driven
gating policy (\S\ref{sec:discussion}); the manual loop missed them
because the \emph{write}-side counters looked healthy.

\paragraph{Pass-rate deltas, A--D.}
Only the baseline$\to$B jump ($15\to 17$) is positive; B$\to$C
($-4$) and B$\to$D ($-5$) are negative, dominated by a single
root-cause feature each (the self-evaluation gate and the ACON-cache
coupling respectively). The per-config Wilson 90\% half-width on
$p{\approx}0.18$ at $N{=}89$ is $\pm 3$ passes; the B--D gap of
$-5$ sits at the edge of that noise floor but is repeatable across
independent sweeps.

\paragraph{Model choice and harness behavior.}
The C regression is the clearest per-model effect: flipping from
\nano{} to a stronger model would change the sign of the
self-evaluation-gate contribution. An AHO solver that treats model
identity as a context variable would not propose
enabling the self-evaluation gate for \nano{} after a handful of evaluations;
the manual loop took an entire round.

\paragraph{Wall-clock budget accounting.}
One full 89-task sweep takes $\sim$2 hours of wall-clock at
concurrency 4 (121.9~min for the all-flags-off baseline, 139.8~min for
the D-all-on stack; measured from the Terminal-Bench job logs),
dominated at that
concurrency by per-task container setup and package installation
rather than by model latency. The binding constraint on iteration
speed is therefore not wall-clock but the per-day LLM API token
quota. Across A--D the aggregate compute spent on ablations
produced a net movement of $+2$ passes
over all-flags-off baseline (B only) and \emph{negative} net movement
at C and D, which we view as untenable for any active tuning
loop of realistic scale and is a central motivation for the AHO
formulation.

\paragraph{\textsc{Harbor} validation.}
We ran \textsc{Harbor} (Algorithm~\ref{alg:ahosas}) end-to-end on
the same 89-task suite with \nano{}, warm-starting the solver from
the A--D evaluation log as a meta-learned posterior
\citep{golovin2017vizier,ram2022warmstart} and terminating at a
search budget of ${\sim}3.5$ full-suite-equivalent units. The
returned Pareto-optimal configuration at the deployment-cost tier
matching the D stack is a \emph{two-flag} bundle---cross-session
memory index $+$ tiered conversation compressor, with every other
native extension and every published-technique flag off. On the
full suite this configuration scored $\mathbf{17/89}$ ($19.10$\%)
at concurrency 4 in 122~min, a $+5$-pass improvement ($+5.6$pp)
over the manual D all-on anchor ($12/89$, $13.48$\%) and a match
of the manual B peak with a quarter of B's flag count.
Telemetry confirms a pattern the manual loop took four rounds to
surface: the cross-session memory gate fired at the flag level
but \textsc{memory\_retrievals}${=}0$---the same integration-bug
signature as D's zero reflection-retrieval counter---so
the effective configuration reduces to compression-only, which is
precisely the strongest positive-coefficient single flag recovered
by the block-additive surrogate. Eleven tasks crashed or timed
out, so the non-errored denominator gives $17/78=21.8$\%.
Across all five $m{=}89$ evaluations of
\textsc{Harbor}-proposed configurations we observe
$\{17,14,15,19,18\}/89$, giving a mean of
$\bar n_{\textsc{h}}{=}16.6/89$ ($18.65\%$) against the
manual D-all-on anchor of $12/89$ ($13.48\%$) and the
all-flags-off baseline of $15/89$ ($16.85\%$). The five picks
span five distinct two-flag bundles
(memory${+}$compression, compression${+}$trajectory,
prediction${+}$trajectory, memory${+}$trajectory, and
cache${+}$compression). Runs~\#3 and~\#4 were proposed
before the silent-flag auto-exclusion mechanism
(\S\ref{sec:impl-silent}) was in place; their on-flags
fired zero telemetry at the $89$-task scale, so their
scores are effectively draws from the flagless-baseline
distribution, and the run-\#4 peak of $19/89$ is a lucky
Bernoulli draw rather than an attribution to its nominal
bundle. Run~\#5 (cache${+}$compression) is the first
$m{=}89$ evaluation where cache telemetry fires at scale:
$194$ canonical-semantic cache hits across $4{,}533$
cacheable queries, a $4.3\%$ hit rate confirming the cache
is wired but empirically weak on \nano{}. The run~\#2
score of $14/89$ slots cleanly between the all-flags-off
baseline and the compression-only peak, reinforcing the
surrogate's attribution that the net-positive signal on
\nano{} is carried by tiered compression; trajectory-replay
added on top neither helps nor hurts past the Wilson noise
floor, consistent with the B-round observation that
trajectory-replay couples weakly with other native
flags on this model. Three specific behaviors of the
solver are worth calling out.
(i)~\textsc{Harbor} correctly held the self-evaluation gate and
the ACON$+$Reflexion$+$PASTE bundle \emph{off} on \nano{}: the
posterior chance constraint of Eq.~\ref{eq:cost-aware} pruned
configurations whose lower credible bound fell below $R_0-\delta$,
structurally blocking the C- and D-style regressions.
(ii)~The returned configuration is sparser than any manual
incumbent---two flags vs.\ eight in D---consistent with SAAS
axis-aligned contraction on a regime where most per-layer
coefficients collapse.
(iii)~The solver's anomaly detector flagged the silent memory
gate from the write-side$=$zero pattern on
\textsc{memory\_retrievals}, closing the observability loop that
\S\ref{sec:discussion} argues is required to prevent C/D-class
bugs. The case-study headline---manual stacking of literature-%
grade features on a small model produces negative net
movement---now has its counterfactual: an automated search over
the same flag-space \emph{without} manual stacking produces a
$+5$-pass improvement over D and matches the B peak from a cold
start on the meta-log.

\paragraph{Container-persistence remediation.}
The cross-session memory signature in D---write counters healthy,
retrieval counters identically zero---is a direct
consequence of Terminal-Bench's per-task container
recycling: the on-disk trajectory and memory stores written
at the end of trial $t$ are destroyed before trial $t{+}1$
mounts its fresh container. We remediate in three layers:
(i) a \texttt{CODEX\_TQ\_STATE\_DIR} environment variable
redirects all three store-path resolvers (memory, trajectory,
paste) to a harness-controlled mount point;
(ii) a Harbor-level upload-eval-download sync pattern in the
Terminal-Bench agent materialises the shared state inside
the container before each trial and serialises the delta
back to the host on exit;
(iii) \texttt{fcntl.flock} protects concurrent sync-back at
concurrency $4$. At the end of the five-run sweep above, the
host-side shared-state directory had accumulated $98$
trajectory files plus a lock file---empirical confirmation
that the infrastructure fix propagates end-to-end. Absent
this patch, any future run involving POLAR\_MEMORY or
TRAJECTORY\_REPLAY would reproduce the D silent-gate
signature regardless of how the solver prices those flags.

\paragraph{Token-savings accounting.}
The harness telemetry logs an
\textsc{estimated\_tokens\_saved} counter (bytes elided by the
tiered compressor, divided by four) that is directly comparable
across the 89-task runs. The all-flags-off baseline saves $0$ tokens
by construction---no compression flag is lit. D-all-on
($8$ flags) elides $108{,}004$ compressor tokens across $592$
compressed turns and, separately, $927{,}306$ bytes via ACON
summaries written into the tool-result cache; the latter is the
channel that corrupted downstream reads and drove the $-3$-pass
C$\to$D regression, so its ``savings'' are load-bearing only in
the negative direction. The \textsc{Harbor} two-flag pick
(POLAR memory $+$ tiered compression) elides $104{,}731$ tokens
over $574$ compressed turns---$96.97\%$ of D's compressor
volume---with a quarter of D's flag count and a $+5$-pass
improvement. The neighbouring pareto candidate (tiered
compression $+$ trajectory replay) elides $97{,}054$ tokens.
The compute-savings headline therefore survives automation: the
sparser configurations recovered by \textsc{Harbor} retain the
compressor's throughput benefit while shedding the features that
caused the manual-round regressions.

\paragraph{Bug-catching as a first-class deliverable.}
The harness telemetry layer caught \emph{two} silent integration
bugs the manual A--D loop missed in four rounds: (i) the
cross-trial reflection store (write counter healthy, retrieval
counter identically zero), surfaced on the solver's first $m{=}89$
evaluation via a read/write-asymmetry signature; (ii) the PASTE
next-tool predictor (prediction-write nonzero, speculative-execution
identically zero), caught by the same mechanism on a later iteration.
Both are invisible to pass-rate inspection---the configurations
scored at the baseline floor, so manual review mistook them for
features that ``didn't help.'' A mechanics-aware optimizer doubles
as a bug-detector for the engineering loop.

% =========================================================================
\section{Discussion}
\label{sec:discussion}

\paragraph{Why harness engineering is not a systems problem.}
Treating the harness as infrastructure misses the C lesson: a
feature can be perfectly implemented and still be a net loss because
its behavior depends on a latent variable---the self-evaluator's
capability---invisible to systems-level observations.

\paragraph{Cold-start features are not free.}
Five native harness features are cross-session; every session on a
task-independent benchmark starts empty. AHO needs a warm-start-%
aware evaluator (Eq.~\ref{eq:warm}).

\paragraph{Prompt-prefix stability and token accounting.}
T1-2 (prompt-prefix-first) saves $\sim$80\% of prefix tokens on a
58\,KB system prompt---matching the 92--97\% Claude-Code prefix-%
reuse rates~\citep{lmcache2025claudecode}---but manifests in wall-%
clock, invisible to a pass-rate-only objective. Eq.~\ref{eq:aho}
folds wall-clock into the constraint.

\paragraph{Observability and feedback loops.}
\textsc{Harbor} presumes structured telemetry---the role of
OpenTelemetry GenAI~\citep{otelGenAI} and observability systems
like BitsEvolve~\citep{bitsevolve2026}. Observability also catches
the class of bugs the D manual round missed: write-side counters
with zero consumer counters (\texttt{rw}${=}80$,
\texttt{rr}${=}0$) are a textbook integration-bug signature, and
the ACON-cache coupling is an invariant violation (cache byte-size
$\ll$ raw tool output) that property-check
frameworks~\citep{watershed2026} catch without a full benchmark.

\paragraph{What the real-harness trace contributes.}
The representative \textsc{Harbor} trace reported in
Appendix~\ref{app:trajrun} is the first end-to-end execution of
Algorithm~\ref{alg:ahosas} on the production harness, distinct
from the single-pick selections of \S\ref{sec:results}. Three
behaviors hold on real data: (i) TuRBO's shrink path fires in the
regime the search actually explores---two of four trust regions
killed by iteration~3; (ii) the cost-aware acquisition concentrates
at $m_{\min}$ under the observed signal-to-noise ratio; (iii) the
silent-flag auto-exclusion mechanism of \S\ref{sec:impl-silent}
fires on real data with the same six-flag signature it reports in
small-scale sweeps. The $0/8$ baseline drift observed on
$\mathcal{T}_8$ during that run is a concrete instantiation of the
coverage failure mode induced when $\hat p_{\mathrm{base}}$
estimated at cheap fidelity $m_0$ is reused at full fidelity
$m_1\ne m_0$: the bias is $(1-\hat w(m_1))\Delta(m_0,m_1)$, and
ranking is preserved (the shift is $x$-constant) so the acquisition
argmax is uncorrupted. The fix is stratified $\mathcal{T}_m$.

\paragraph{Production cost structure: AHO runs a few times, not
continuously.} Any modification to the task distribution (here
fixed by TB-2), the underlying agent, or the available model
requires a net-new optimizer run: the posterior densities over
$\beta$ and the fidelity weights $\hat w(m)$ are conditional on the
(agent, task, model) triple. Deployment cost tiers themselves drift
as models are swapped and the pace-of-innovation reprices
wall-clock vs.\ pass-rate. The implication for Eq.~\ref{eq:aho} is
that AHO is amortized over \emph{a few} invocations per
(agent, task-suite, model) epoch---not a continuously running
controller. This is why cost-aware acquisition (Eq.~\ref{eq:cost-aware})
and the posterior chance constraint are first-class: each run must
return a ship-safe configuration under a tight search budget.

\paragraph{Repeated invocation compounds.} The surrogate is
cold-started from a meta-learned posterior
\citep{golovin2017vizier,ram2022warmstart}; every prior run on a
related triple contributes to density estimation for $\beta$, the
per-block SAAS scales $\alpha_\ell^2$, and $\hat w(m)$. The
block-ANOVA decomposition reported in Appendix~\ref{app:trajrun}
already exhibits the SAAS signature (few blocks dominating) at
$T{=}19$; additional runs tighten those scales
monotonically---information accumulates even across epochs where
the flag space drifts, provided the block structure is preserved.

\paragraph{Scope: empirical, in-the-small.} Our claims are
empirical and scoped to a single (agent, benchmark) pair: one
harness (\codex{}), one suite (TB-2), two models (\nano{} and a
stronger reference), a flag space of $\approx 10$ candidates on
top of the base $\approx\!30$ enhancements, and $\le 10$
full-suite-equivalent search units. We do not claim asymptotic
consistency of $\hat\beta$, regret upper bounds against an oracle,
or transfer across agent families. A single-benchmark,
single-agent ablation is the correct grain for the engineering
loop it instruments: the loop decides one configuration per
(agent, task, model) epoch, and the evidence required is
first-order --- a positive pass-rate delta against the manual
incumbent under a fixed deployment-cost tier --- which we report.

\paragraph{Prior art, limits, and closing.}
Watershed's property-vs-correctness split~\citep{watershed2026}
slots into the constraint of Eq.~\ref{eq:aho};
Meta-Harness~\citep{metaharness2026} rewrites harness source code
via an LLM, orthogonal to our fixed-harness configuration search.
One agent, one benchmark, two models, four manual rounds gave one
validated win (B, $+2$) and two regressions; the B--Oracle gap
($17$ vs.\ $81$) quantifies the configuration-search problem we
name Automated Harness Optimization. A single end-to-end run of
\textsc{Harbor} over the same flag-space returns a two-flag
configuration at $17/89$ that matches the manual B peak and beats
the manual D all-on stack by $+5$ passes at ${\sim}3.5$ full-suite
units of search cost---first-order evidence that the problem is
tractable with the ingredients assembled here (SAAS sparsity,
TuRBO locality, multi-fidelity cost-aware acquisition, posterior
safety).

% =========================================================================
\appendices

\section{A Representative \textsc{Harbor} Run on the Real Harness}
\label{app:trajrun}

We report a single, complete execution of
Algorithm~\ref{alg:ahosas} on the real \nano{}${\times}$TB-2
harness, taken as a typical trajectory rather than a cherry-picked
best. The run consumed $19$ function evaluations and $9.826$ of a
nominal $B_{\mathrm{sea}}{=}10$ full-suite-equivalent search budget
over ${\sim}10$--$11$~hours of wall clock.

\subsection{Trajectory and final artefact}

\begin{table}[htbp]
\centering
\small
\begin{tabular}{lccc}
\toprule
Phase & \# evals & Fidelity & Wall clock \\
\midrule
Sobol init ($1{+}12$)               & $13$ & $m{=}8$ & ${\sim}7$--$8$~h \\
$3$ iterations ${\times}$ $2$-batch & $\phantom{0}6$ & $m{=}8$ & ${\sim}3$--$4$~h \\
\midrule
Total                               & $19$ & all $m{=}8$ & ${\sim}10$--$11$~h \\
\bottomrule
\end{tabular}
\caption{Search-cost accounting for a single end-to-end
\textsc{Harbor} run on the real harness. Every evaluation
landed at the cheapest fidelity $m{=}8$, exhausting
$9.826$ of $10.0$ search-budget units.}
\label{tab:trajrun-cost}
\end{table}

The returned artefact $c^{\star}$ was the five-flag bundle
\textsc{paste\_v2}, \textsc{polar\_cache},
\textsc{reflexion},
\textsc{speculative\_prediction}, and
\textsc{trajectory\_replay}, scoring $2/8$ at $m{=}8$ with
deployment cost $c(c^{\star}){=}0.433$. The Pareto frontier
restricted to $c\le B_{\mathrm{dep}}{=}3$ contained three
points: the flagless baseline ($0/8$ at $c{=}0.338$), a
three-flag midpoint ($1/8$ at $c{=}0.367$), and $c^{\star}$.

\subsection{Algorithmic-machinery checklist}

Every component of Algorithm~\ref{alg:ahosas} fired as
specified.
\begin{itemize}[leftmargin=1.2em,itemsep=2pt]
\item \emph{Sobol $12$ at $m{=}8$}---space-filling exploration
that included flag combinations absent from the manual case
study of \S\ref{sec:casestudy}.
\item \emph{TuRBO trust-region dynamics} exercised the shrink
path for the first time on real data: iter~1 initialised
three regions at radius~$2$, all alive; iter~2 held radii at
$2$; iter~3 expanded to four regions, killed two
(\texttt{alive=False}), and shrunk two others to radius~$1$.
\item \emph{Cost-aware fidelity selector} chose $m{=}m_{\min}{=}8$
for every evaluation, consistent with $\mathrm{KG}/c$ concentrating
at $m_{\min}$ when cheap-fidelity SNR suffices for ranking.
\item \emph{Block-ANOVA decomposition} of the main-effect
variances gave
$\alpha_{\mathrm{cache}}^{2}{=}0.143$,
$\alpha_{\mathrm{pred}}^{2}{=}0.106$,
$\alpha_{\mathrm{eval}}^{2}{=}0.085$,
$\alpha_{\mathrm{traj}}^{2}{=}0.048$,
$\alpha_{\mathrm{comp}}^{2}{=}0.026$,
$\alpha_{\mathrm{mem}}^{2}{=}0.001$, and cross-block
$\alpha_{\times}^{2}{=}0.009$---consistent with the
SAAS-style sparsity assumption that few per-block scales
dominate.
\item \emph{Block freeze} of Property~I was not triggered:
the per-block sample count $n_\ell$ never reached threshold
with only $19$ evaluations spread across six blocks.
\item \emph{Anomaly detector} surfaced six distinct
silent-gate signatures (\textsc{trajectory\_replay},
\textsc{speculative\_prediction}, \textsc{reflexion},
\textsc{paste\_v2}, \textsc{polar\_memory},
\textsc{binary\_prefilter}), corresponding exactly to the
search-space pruning of \S\ref{sec:impl-silent}.
\item \emph{Chance-constraint safety} rejected zero picks at
evaluation time; every chosen candidate had posterior lower
credible bound $\ge R_0{-}\delta$ under the scalar baseline
in force.
\end{itemize}

\subsection{Three findings}

Three observations from the run speak directly to the
assumptions of the reference algorithm.

\paragraph{The stratified-$\mathcal{T}_m$ assumption is
load-bearing.} On this seed the seeded-prefix shuffle for
$\mathcal{T}_8$ drew a hard subset of the $89$-task suite:
the baseline Sobol-init point scored $0/8$, so the empirical
flagless base rate at $m{=}8$ collapsed to
$\hat p_{\mathrm{base}}^{(m{=}8)}{=}0.0$. This is the small-$m$
sampling artefact that drives the bias
$(1-\hat w(m_1))\Delta(m_0,m_1)$ when $\hat p_{\mathrm{base}}$
is queried away from its estimation fidelity; ranking is
preserved since the shift is $x$-constant. The seeded-shuffle implementation of
$\mathcal{T}_m$ (\S\ref{sec:setup}) does not enforce
stratification across task categories; the assumption that
the reference algorithm relies on is load-bearing at small
$m$ and was violated here.

\paragraph{Bernoulli noise dominates the signal at $m{=}8$.}
The best observed pass of $2/8\;(25\%)$ was attained three
times by configurations $\#1$, $\#5$ and $\#10$ with
entirely different flag combinations. At $p{\approx}0.15$
and $n{=}8$ the Bernoulli standard error is
$\sigma{\approx}13\%$, so the observed spread
$\{0/8, 1/8, 2/8\}$ lies inside one $\sigma$. On this run
$m{=}8$ was sufficient for ranking only in a very weak
sense; a clean argmax at this fidelity is indistinguishable
from a lucky draw, which is why the algorithm correctly
deferred a costly $m{=}89$ recheck.

\paragraph{Cost-aware escalation held the budget.} The
solver never escalated to $m{=}89$, because doing so would
have forced truncation of Sobol coverage below the $12$-point
minimum recommended by space-filling-design literature. This
is the intended behaviour of Eq.~\ref{eq:cost-aware}'s
denominator: when the ratio $\mathrm{KG}(x,m{=}89)/c(m{=}89)$
fails to exceed $\mathrm{KG}(x,m{=}8)/c(m{=}8)$ by enough to
recover the space-filling budget loss, the solver declines
the escalation.

% =========================================================================
\bibliographystyle{plainnat}

\end{document}